# Genetic Algorithm based hyper-parameters optimization for transfer Convolutional Neural Network

Chen Li[1,2], JinZhe Jiang[1,2], YaQian Zhao[1,2], RenGang Li[1,2], EnDong Wang[1,2], Xin Zhang[3,4,5]*, Kun Zhao[6]

1. State Key Laboratory of High-end Server & Storage Technology, Beijing, China, 100085
2. Inspur Group Co., Ltd, Beijing, China, 100085
3. Shandong Hailiang Information Technology Institutes, Jinan, China, 250014
4. State Key Laboratory of High-End Server & Storage Technology, Jinan, China, 250014
5. Inspur Electronic Information Industry Co., Ltd, Jinan, China, 250014
6. Guangdong Inspur Intelligent Computing Technology Co., Ltd, Guangzhou, China, 510000
* xzphys@gmail.com

## Abstract

Hyperparameter optimization is a challenging problem in developing deep neural networks. Decision of transfer layers and trainable layers is a major task for design of the transfer convolutional neural networks (CNN). Conventional transfer CNN models are usually manually designed based on intuition. In this paper, a genetic algorithm is applied to select trainable layers of the transfer model. The filter criterion is constructed by accuracy and the counts of the trainable layers. The results show that the method is competent in this task. The system will converge with a precision of 97% in the classification of Cats and Dogs datasets, in no more than 15 generations. Moreover, backward inference according the results of the genetic algorithm shows that our method can capture the gradient features in network layers, which plays a part on understanding of the transfer AI models.

## 1. Introduction

Convolutional neural networks (CNN) now is an extensively used artificial intelligence model in computer vision tasks [1]. However, a great deal of labeled data is required for training process, which sometimes is not easy to obtain. Also, it is inefficient to restart training a CNN model from the very beginning on every task. Transfer learning can be used in these situation to improve a model from one domain to another related one by transferring information.

Oquab et al. come up with a method to transfer a pre-parameterized CNN model to a new task [2]. With this pre-trained parameters, they fine-tune the original model to a target model. The only difference is that an additional network layer is added to the pre-parameterized model. To adapt the target dataset, the additional layer is fine-tuned from the new task with small samples.

With lots of refined datasets established, it is reasonable to use ready-made datasets as a reference and take this advantage to a fresh task. To date, transfer learning has become a widely used technique in many area successfully, such as text sentiment classification [3], human activity classification [4], image classification [5]-[7], and multi-language text classification [8]-[10].

Transfer learning technique has been widely used to solve challenging applications and has shown its potential, while the mechanism behind is still ambiguous. Just like the clouds of deep neuron networks, interpretability is one of the challenging questions for transfer learning. Especially in the case of transfer of CNN, it is difficult to design the hyper-parameter, for instance, which layers should be trainable and which frozen because of the interpretability problem. So far, all of these are based on manual design. However, the parameter space increases exponentially or sub-exponentially with the NN layers, which makes it difficult to find an optimized solution by trial and error.

In this paper, an automatically learning the hyper-parameters method is proposed on a transfer CNN model. Only one hyper-parameter, the trainability of parameters in layers, is considered in this work. Under this condition, the search space has the exponential relationship with the number of layers. Instead of ergodic search, we adopt the genetic algorithm (GA) to explore the trainability of CNN layers.

The GA constructs an initial population of individuals, each individual corresponding to a certain solution. After genetic operations performed, the population is pushed towards the target we set. In this paper, the state of all the layers are encoded as a binary string to represent the trainability of networks. And selection, mutation and crossover are defined to imitate evolution of population, so that individual diversity can be generated. After a period of time, the excellent individuals will survive, and the weak ones will be terminated. To quantify the quality of individuals, the accuracy of the CNN model and the number of trainable layers are adopted, which embodies in the form of the fitness function. For each individual, we perform a conventional training process, including the techniques that are widely used in deep learning field. And for the whole population that is consist of individuals in the same generation, the genetic operations are performed. The process ends up with the stop criterion reaches.

As it needs to carry through a whole CNN training process in the all population, the genetic process is computationally expensive. In view of this, several small datasets (cats_vs_dogs, horses or humans and rock_paper_scissors) [11]-[13] are selected to test the genetic process. Here, we demonstrate the ability of the GA to search key layers to be fixed (or to be trained). And then the implication of important layers is analyzed to make a further understanding of the models. The GA shows a robust result to obtain the best transfer model.

The following of this paper is organized into 4 sections. First, Section 2 introduces the related work. And in Section 3, we briefly illustrate the details of the GA to search the space of the transfer model's trainability. Section 4 gives the experiment results. And conclusions are drawn in Section 5.

## 2. Related Work

Our method is related to the works on CNN, transfer learning, and the GA on hyper-parameter optimization, which we briefly discuss below.

**Convolutional Neural Networks.** A neural network is a network connected by artificial nodes (or neurons). The neurons are connected by tunable weights. And an activation function controls the amplitude of the output. Neural networks are verified to be capable of recognition tasks [14]. CNN is a particular neural network with a hierarchical structure. The convolution operation is carried out in specific neurons that are adjoining in spatial. In the general model, assume layer $p$ give outputs $A$, and this output $A$ will then convoluted with a filter to transport the information to the layer $(p+1)$. The activation function is performed then to define the outputs. During the training process, error signals are computed back-propagating the CNN model. Here, error is calculated by a certain way according to the difference between the supervision and prediction of the outputs. In the past years, the establishing of large-scale datasets (e.g., ImageNet [15]) and the advance of hardware make it possible to train deep CNN [16][17] which significantly outperform Bag-of-Visual-Words [18]-[20] and compositional models [21]. Recently, several efficient methods were combined with the CNN model, such as ReLU activation [16], batch normalization [22], Dropout [23] and so on. With the assistance of methods mentioned above, the CNNs [16][17] have shown the state-of-the-art of the performance over the conventional method [18]-[21] in the area of computer vision.

**Transfer learning.** Transfer learning is a method that aims to transfer experience or knowledge from original source to new domains [24]. In computer vision, two examples of transfer learning attempt to overcome the shortage of samples [25],[26]. They use the classifiers trained for task A as a pre-trained model, to transfer to new classification task B. Some methods discuss different scene of transfer learning, which the original domains and target domains can be classified into the same categories with different sample distributions [27]-[29]. For instance, same objects in different background, lighting intensity and view-point variations lead to different data distributions. Oquab et al. [2] propose a method to transfer a pre-parameterized CNN model. In their work, they show that the pre-trained information can be reused in the new task with a fairly high precision. This transfer CNN model carry out the new task successfully, also save the training time passingly. Some other works also propose transferring image representations to several image recognition tasks, for instance image classification of the Caltech256 dataset [30], scene classification [31], object localization [32],[33], etc. Transfer learning is supposed to be a potential approach.

**Genetic algorithm on hyper-parameter optimization.** The genetic algorithm is a kind of a heuristic algorithm inspired by the theory of evolution. It is widely used in search problems and optimization problems [34],[35]. By performing biological heuristic operators such as selection, mutation and crossover. The GA becomes a useful tool in many areas [34]-[41].

A standard GA translates the solution of the target problem into codes, then a fitness function is constructed to evaluate the competitiveness of individuals. A typical example is the travelling-salesman problem (TSP) [36], which is a classical NP-hard problem in combinatorial optimization on optimizing the Hamiltonian path in an undirected weighted graph.

A GA can generate various individual genes, and can make the population evolved in the genetic process. Selection, mutation and crossover are common methods of genetic process. The selection process imitate natural selection to select the superior and eliminate the inferior. Mutation and crossover process makes it possible to produce new individuals. The specific technical details of mutation and crossover operations are usually based on the specific tasks. For instance, mutation operation can be designed to flip a single bit for binary encoding.

Some previous works have already applied the GA to learning the structure [37][38] or weights [39][40] of artificial neural networks. Xie et al. [41] optimize the architectures of CNN by using the GA. The idea of their work is that encoding network state to a fixed-length binary string. Subsequently, populations are generated according the binary string. And every individual is trained on a reference dataset. Then evaluating all of them and performing the selection process and so on. They perform the GA on CIFAR-10 dataset, and find that the generated structures show fairly good performance. These structures are able to employ for a larger scale image recognition task than CIFAR-10 such as the ILSVRC2012 dataset.

Suganuma et al. [42] apply Cartesian genetic programming encoding method to optimize CNN architectures automatically for vision classification. They construct a node functions in Cartesian genetic programming including tensor concatenation modules and convolutional blocks. The recognition accuracy is set as the target of Cartesian genetic programming. The connectivity of the Cartesian genetic programming and the CNN architecture are optimized. In their work, CNN architectures are constructed to validate the method using the reference dataset CIFAR-10. By the validation, their method is proved to be capable to construct a CNN model that comparable with state-of-the-art models.

The GA is applied to solve the hyper-parameter optimization problem in another work proposed by Han et al. [43]. In [43], the validation accuracy and the verification time are combined into the fitness function. The model is simplified to a single convolution layer and a single fully connected layer. They evaluated their method with two datasets, the MNIST dataset and the motor fault diagnosis dataset. They show the method can make the both the accuracy and the efficiency considered.

Young et al. [44] propose a GA based method to select network on multi-node clusters. They test the GA to optimize the hyper-parameter of a 3-layer CNN. The distributed GA can speed up the hyper-parameter searching process significantly. Real et al. [45] come up with a mutation only evolutionary algorithm. The deep learning model grows gradually to find a satisfactory set of combinations. The evolutionary process is slow due to the mutation only nature. Xiao et al. propose a variable length GA to optimize the hyper-parameters in CNNs [46]. In their work, they does not restrain the depth of the model. Experimental results show they can find satisfactory hyper-parameter combinations efficiently.

# 3. Method

In this section, we introduce the method of GA for learning the trainable layers of transfer CNN. In general, the state of all the layers are encoded as a binary string to represent the trainability of networks. Following, selection, mutation and crossover are defined to imitate evolution of population, so that individual diversity can be generated and excellent characters can be filtrate out.

Throughout this work, the GA is adopted to explore the trainability of the hidden layers. The network model, optimizer, base learning rate and other hyper-parameters of each individual are obtained via an empirical selection and are not optimized specifically.

## 3.1 Details of Genetic Algorithm

Considering the states of the networks, each layer has two possibilities, trainable or frozen, so a T layers network will give $2^T$ possible states. Due to the difficulty of searching an exponential space, we simplify the problem on the case that the labels of trainable layers are continuous, which means the state of the model should be a sandwich-shape (Frozen_layers-Trainable_layers-Frozen_layers, shown in Figure 1). Then the tunable parameters can be set to the label of the start layer $L_s$ and the label of the end layer $L_e$. That makes a bivariate optimization problem, which will change the $2^T$ space to T×(T-1)/2. The flowchart of the genetic process is shown in Algorithm 1.

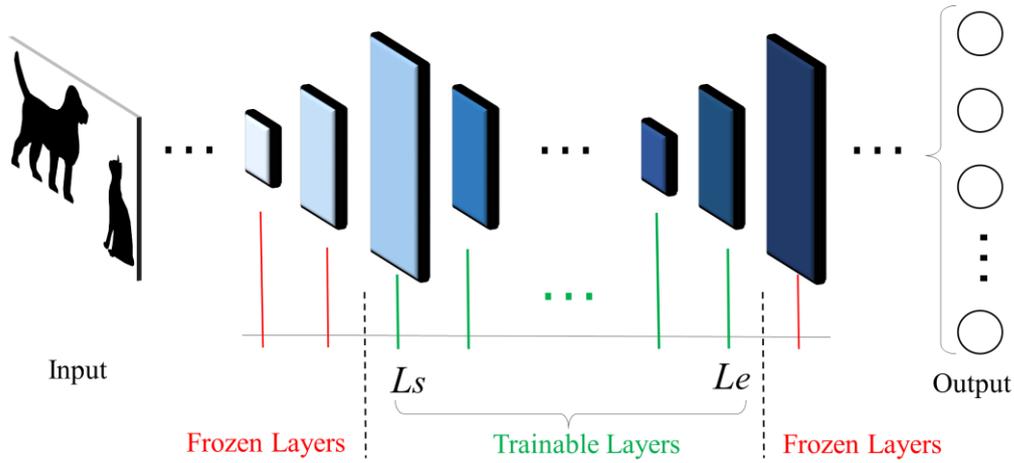

Figure 1. Transfer CNN model in sandwich-shape encoding, Ls and Le are tunable parameters to determine the boundary of trainable layers

The GA is performed by N generations, and very round between generations consists of selection, mutation and crossover process, respectively.

**Initialization.** The population is set to M individuals (M=50 in our case). And the genes for each individual is initialized in a binary string with D bits (In our case, the bounds of two parameters $L_s$ and $L_e$ are 0 and 156, respectively. To represent the number with the length of 156 by a binary string, 8 bits are needed because $2^8$=256. And consider that there are two parameters, the total bits of D is then set to 8*2=16). Here, all the bits are randomized to either 0 or 1, independently.

**Mutation and Crossover.** As the bits are set to a binary encoding, for each individual with D bits, the mutation process involves flipping every bit with the probability $q_M$. The set of $q_M$ will affect the exploration breadth and the rate of convergence. Instead of randomly choosing every bit individually, the crossover process consider exchange fragments of two individuals. Here the fragments are the subsets in individuals, for purpose of hold the useful characters in the form of binary schema. Each pair of corresponding fragments are exchanged with the probability $q_C$ (0.2 in our case).

**Evaluation and Selection.** In this paper, the selection process is performed after mutation and crossover. A fitness function F is used to identify excellent individuals, which is defined as the Eq. 1:

$$F_{i,j} = acc_{i,j} - \gamma \cdot (L_e - L_s) \tag{Eq. 1}$$

where $acc_{i,j}$ is the accuracy for the j-th individual in the i-th generation obtained from testing of the CNN model. γ is the weight of layer number (0.005 in our case). Although it is not necessary that the more the trainable layers open, the better accuracy the model will be (details shown in section 4.1), we introduce the number of trainable layers as a part of component of fitness function.

Fitness impacts the probability that whether the j-th individual is selected to survive. A Russian roulette process is performed following the Eq. 2 to determine which individuals to select.

$$P_{i,j} = \frac{F_{i,j}}{\sum_{j=1}^{M} F_{i,j}} \tag{Eq. 2}$$

where $P_{i,j}$ is the probability for j-th individual to survive. According to the Russian roulette process, the larger the fitness value of individual is, the more probable the individual will survive.

---

**Algorithm 1** The Genetic Algorithm for Trainable Layers Decision

1. **Input:** the dataset I, the pre-trained model P, the number of generations N, the number of individuals in each generation M, the mutation parameter $q_M$, the crossover parameter $q_C$, and the weight of layer number.

2. **Initialization:** the genes for each individual is initialized in a binary string with D bits, all the bits are randomized to either 0 or 1. Performing training process to get accuracies of each individuals;

3. **for** t = 1, 2, . . . , N do

4.     **Crossover:** for each pair, performing crossover with probability $q_C$;

5.     **Mutation:** for each individuals, performing mutation with probability $q_M$;

6.  **Selection:** producing a new generation with a Russian roulette process;

7.  **Evaluation:** performing training process to get accuracies of for the new population. And check the convergence, jump out of the loop if the stopping criterion is satisfied;

8. **end for**

9. **Output:** M individuals in the last generation with their recognition accuracies.

### 3.2 Details of transfer CNN model

The MobileNetV2 model developed by Google [47],[48] are used as the base model in our case. This model is pre-trained on the ImageNet dataset [15], which consisting of 1.4 M images and can be classified into 1000 categories. In this work, this base of knowledge is used to be transferred to classify specific categories with different datasets.

In the feature extraction experiment, one way to design a transfer CNN model is adding layers on top of the original model. Then the original model is fixed with the structure and some of the weights. And the rest part is trained to transfer toward the specific classification problem. During the process, the generic feature maps is retained, while the changing weights and the adding layers are optimized to specific features. Besides the top layers, the performance can be even further improvement by fine-tune the parameters of other layers of the pre-trained model, which is usually an empirical process. In most convolutional networks, it is believed that the early layers of the model learn generic features of images, such as edges, textures, etc. With the layers forward to the tail layers of the model, the features extracting by CNN become more specific to the target domain. The goal of transfer learning is to preserve the generic parts of the model and update the specialized features to adapt with the target domain.

In this work, the task is simplified to transfer the MobileNetV2 pre-trained model on several classification problems. Instead of manual adjustment, the GA is used to optimize the trainability of the hidden layers of the transfer model.

## 4. Experiments

ASIRRA (Animal Species Image Recognition for Restricting Access) is a Human Interactive Proof that works by asking users to identify photographs of animals. They've provided by Microsoft Research with over three million images of cats and dogs [11] (Dataset 1). For transfer learning, we use 0.1% for training and 99.9% for testing. Horses or Humans is a dataset of 300×300 images in 24-bit color, created by Laurence Moroney [12] (Dataset 2). The set contains 500 rendered images of various species of horse and 527 rendered images of humans in various poses and locations. Rock Paper Scissors is a dataset containing 2,892 images of diverse hands in Rock/Paper/Scissors poses [13] (Dataset 3). Each image is 300×300 pixels in 24-bit color. We use 10% for training and 90% for testing.

## 4.1 Verification of the trainable layers effect

We verify the impact of different trainable layer by checking the recognition accuracy on the Dataset 1 [11]. It should be noted that some of the layers don't have any parameters to be trained originate from the MobileNetV2 structures [47]. For instance, layer 154 and 155 have no parameters, while the case open 150-155 as trainable layers is still shown in the figure 1 for a rough comparison.

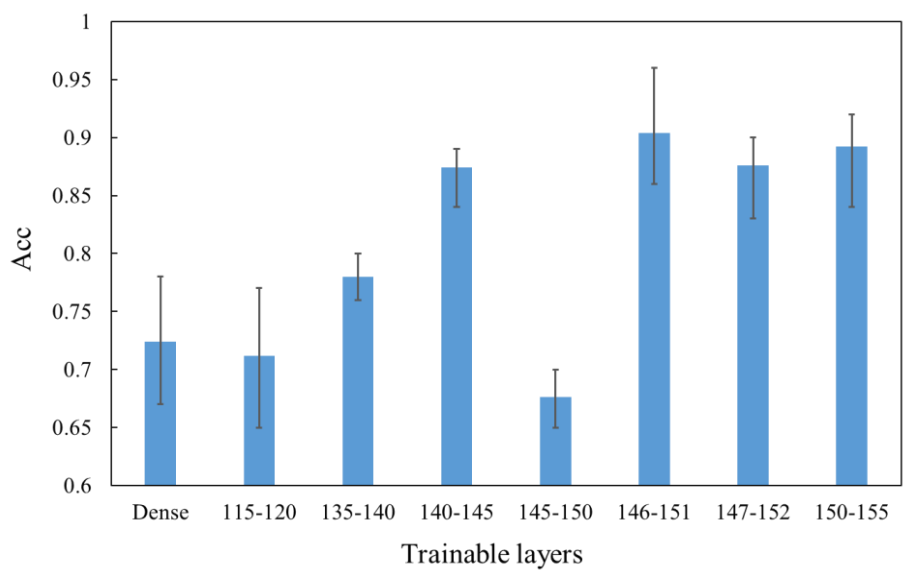

Figure 2. The rough comparison of accuracy and different trainable layers. The legend below means open different layers to be trainable, "dense" means only the top fully connected layer is trainable. Each case is test for 20 rounds to get the average accuracy.

Fig. 2 shows a conflict with the intuition we introduced in Section 3.2. Training the layers 135-140 makes the accuracy higher than the layers of 115-120, while training the layers 147-152 decrease the accuracy compared with the case of 146-151. That means with the layers higher up, the features are not necessary to be more specific to the target dataset. Also, the result indicates the choice of trainable layers is of vital importance.

## 4.2 Optimization result from genetic algorithm

To verify the performance of GA on the transfer CNN tasks, three datasets (Dataset 1, 2 and 3) are tested. The result is shown in Fig. 3. With the genetic operations, it shows a significant improvement in the average accuracy on all the three datasets. Especially for the Dataset 3, the accuracies in the first generation are barely better than a random choice. While, after the system converged, the best individual achieves the accuracy of 97%. At around the 14th generation, the system is converged and gives the average recognition accuracies at 93%, 90% and 87% of the three dataset, respectively.

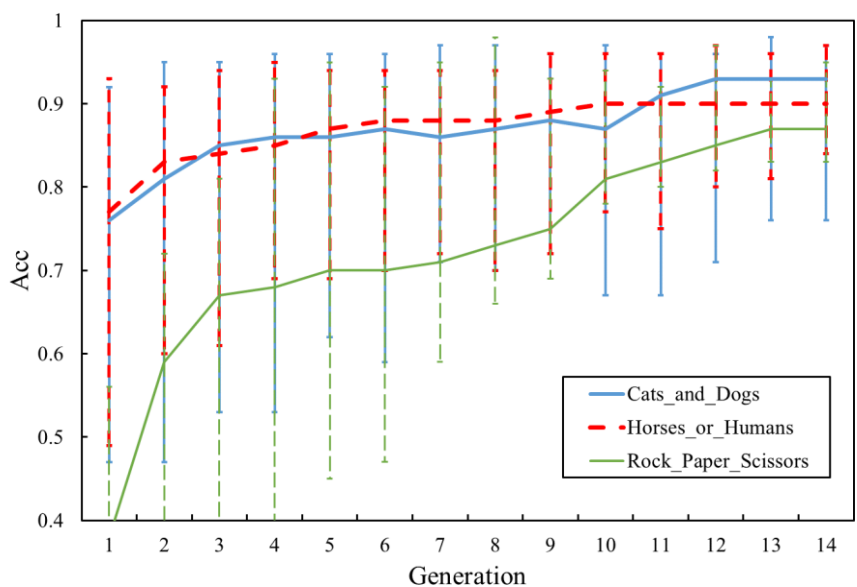

Figure 3. The average accuracy over all individuals with respect to the generation number. Blue solid line, red solid line and green dash line correspond to three datasets, respectively. The bars indicate the highest and lowest accuracies in the corresponding generation.

The results of Dataset 1 are summarized in Table 1. The average recognition accuracy is updated from 76% to 88% by generation. The best individuals and the worst individuals are also improved with the genetic process. Although there is a fortunate fluke that the best individual gives a fairly high accuracy in the first generation, it still can be proved that the GA is more efficiency than random search. For the Dataset 2 and 3, see the SI (Supplementary Information, Table 2 and Table 3).

Table 1. The Recognition accuracy on the Cats and Dogs dataset testing set. The best individual in corresponding generation is translated to trainable layer numbers.

| Gen | Max | Min | Avg | Start Layers | End Layers |
| --- | --- | --- | --- | --- | --- |
| 1 | 0.92 | 0.47 | 0.76 | 131 | 133 |
| 2 | 0.95 | 0.47 | 0.81 | 147 | 151 |
| 3 | 0.95 | 0.53 | 0.85 | 130 | 155 |
| 4 | 0.96 | 0.53 | 0.86 | 130 | 155 |
| 5 | 0.96 | 0.62 | 0.86 | 123 | 151 |
| 6 | 0.96 | 0.59 | 0.87 | 127 | 151 |
| 7 | 0.97 | 0.71 | 0.86 | 146 | 151 |
| 8 | 0.97 | 0.70 | 0.87 | 124 | 151 |
| 9 | 0.96 | 0.72 | 0.88 | 147 | 151 |
| 10 | 0.97 | 0.67 | 0.87 | 124 | 151 |
| 11 | 0.96 | 0.67 | 0.88 | 129 | 151 |
| 12 | 0.96 | 0.71 | 0.90 | 129 | 151 |
| 13 | 0.98 | 0.76 | 0.91 | 129 | 151 |
| 14 | 0.97 | 0.76 | 0.91 | 129 | 151 |

## 4.3 Characterization of neural network layers

After being translated to trainable layer numbers, the best individuals in each generations are shown in Table 1. For the case of Dataset 1, the result converged to the 129-151 as trainable layers. While different Datasets will give specific selection of trainable layers (142-151 for Dataset 2, 130-136 for Dataset 3. Shown in SI). It reveals that even with the same model, the importance of network layers is different for different tasks. That maybe originates from that the specialized features are composed of generic features extracting by different network layers. So, different datasets correspond to specialized features, which corresponding to specific network layers.

To investigate the responding of the network layers, the gradients information is then analyzed. Figure 4 shows the result of the maximum value of gradients in each layers activated by dogs images and cats images, respectively. It shows the maximum gradients of nodes in each layer are not sensitive to different categories in Dataset 1.

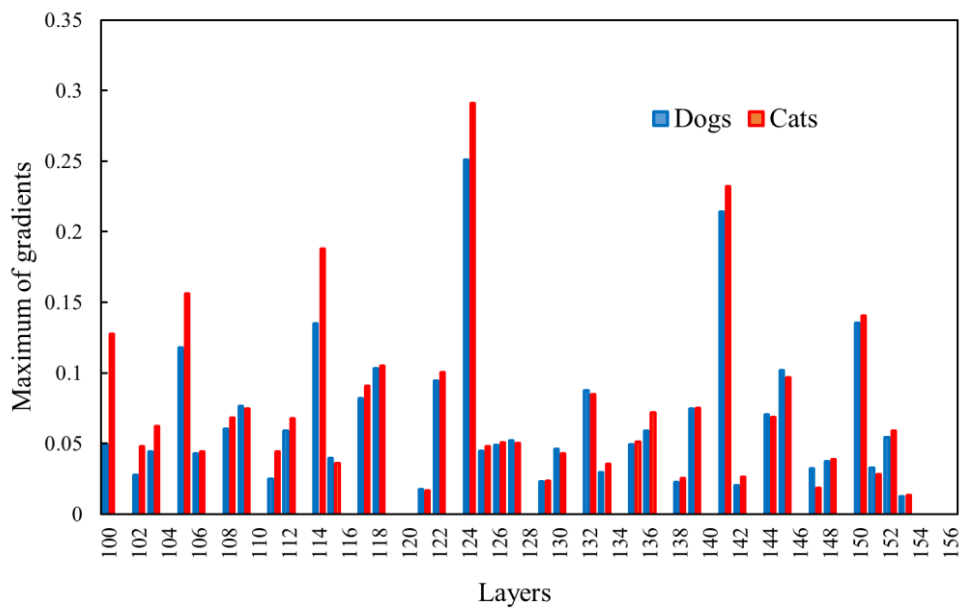

Figure 4. The maximum value of gradients in each layer activated by dogs/cats images

Figure 5(a) shows that it is distinguishable of two categories by the average value of gradients in the layers, but with a complicated features. It is worth emphasizing that this comes from an average of the dataset ensemble, which is maybe not necessarily consistent with a single sample. While summation of gradients reveals some features are distinctly different between two categories, which is shown in Figure 5(b).

The summation of gradients in the layer 105, 114, 124, 132, 141 and 150 are significantly higher than others. For the layer 114, 132 and 150, the sign of summation value of two categories are even opposite on average, which maybe contribute to the classification as a criterion. It is interesting that the layer 150 is also the boundary of the optimized trainable layers by GA. That inspires us with an explainable AI perspective, although it is still superficial. However, the start layers optimized by GA is converged to 129, which is affected by the weight of layer number γ. It still need further works on the effect of this factor.

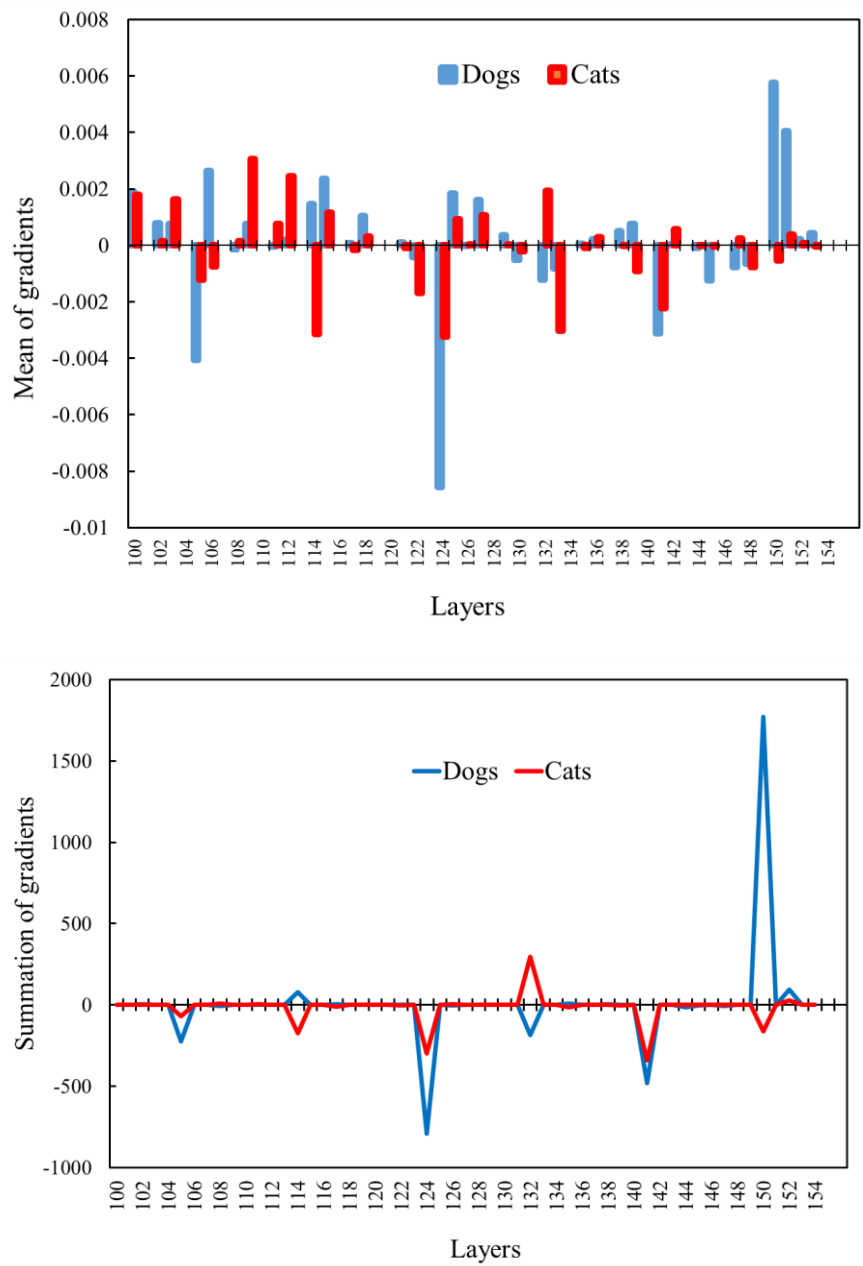

Figure 5. (a) The average gradient of each layers activated by dogs/cats images (b) Summation of gradients by all the nodes in the same layer

## 5. Conclusions

In this paper, we apply the GA to learn to decide the trainable layers of transfer CNN automatically. Our main idea is to encode the trainable layers number as a gene of individuals, and update the population by genetic operations to obtain the best transfer CNN networks. We perform the GA on three datasets (cats_vs_dogs, horses or humans and rock_paper_scissors). The results demonstrate the availability of the GA to apply to this task.

Moreover, according this GA guided results, we can acquire more information by analyzing other features such as gradients. This backward inference can help us understanding the transfer AI models.

Although we find some essential information from the analysis of gradients, it is challenging to interpret AI models by the information so far, even to give an insight of design the transfer CNN. However, it's an open question for the interpretability of AI model. Our approach may help to this goal. Further analysis can help us learn more from AI models, help us moving on towards explainable AI models.

DNA computing, as an alternative technique of computing architecture, uses DNA molecules to store information, and uses molecular interaction to process computing [49]. The parallelism is the advantage of DNA computing compared with electronic computer, which can speed up exponentially in some cases. The GA can be implemented by DNA computing naturally. With the DNA computing based GA, it may greatly speed up hyper-parameter optimization process in future.

## Acknowledgements

This work was supported by National Key R&D Program of China under grant no. 2017YFB1001700, and Natural Science Foundation of Shandong Province of China under grant no. ZR2018BF011.